\documentclass{article}

\usepackage[final]{nips_2017}

\usepackage[utf8]{inputenc} 
\usepackage[T1]{fontenc}    
\usepackage{hyperref}       
\usepackage{url}            
\usepackage{booktabs}       
\usepackage{amsfonts}       
\usepackage{nicefrac}       
\usepackage{microtype}      

\usepackage{amsmath,amsfonts,amssymb,amsthm}
\usepackage{bm}
\usepackage{graphicx}

\newtheorem{assumption}{Assumption}

\newcommand{\EE}[1]{\mathbb{E}\left[ #1 \right]}
\newcommand{\1}[1]{\ensuremath{ \mathbb{I}_{(#1)} }}
\newcommand{\T}{\ensuremath{ \top }}

\newcommand{\ind}{\ensuremath{ \,\, \bot \,\, }}
\newcommand{\given}{\ensuremath{ \mid }}
\newcommand{\R}{\ensuremath{\mathbb{R}}}

\renewcommand{\d}[1]{\ensuremath{\operatorname{d}\!{#1}}}
\newcommand{\set}[1]{\left\{ #1 \right\}}

\newcommand{\vect}[1]{\bm{\mathrm{#1}}}

\newcommand{\matr}[1]{\mathrm{#1}}

\newcommand{\riskpapers}{li2015physiological,schulam2015framework,alaa2016personalized,wiens2016patient,cheng2017sparse}

\title{Reliable Decision Support using\\Counterfactual Models}

\author{
  Peter Schulam \\
  Department of Computer Science \\
  Johns Hopkins University \\
  Baltimore, MD 21211 \\
  \texttt{pschulam@cs.jhu.edu}
  \And
  Suchi Saria \\
  Department of Computer Science \\
  Johns Hopkins University \\
  Baltimore, MD 21211 \\
  \texttt{ssaria@cs.jhu.edu}  
}

\begin{document}

\maketitle

\begin{abstract}
  Decision-makers are faced with the challenge of estimating what is likely to
  happen when they take an action. For instance, if I choose not to treat this
  patient, are they likely to die? Practitioners commonly use supervised
  learning algorithms to fit predictive models that help decision-makers reason
  about likely future outcomes, but we show that this approach is unreliable,
  and sometimes even dangerous. The key issue is that supervised learning
  algorithms are highly sensitive to the policy used to choose actions in the
  training data, which causes the model to capture relationships that do not
  generalize. We propose using a different learning objective that predicts
  \emph{counterfactuals} instead of predicting outcomes under an existing action
  policy as in supervised learning. To support decision-making in temporal
  settings, we introduce the Counterfactual Gaussian Process (CGP) to predict
  the counterfactual future progression of continuous-time trajectories under
  sequences of future actions. We demonstrate the benefits of the CGP on two
  important decision-support tasks: risk prediction and ``what~if?''  reasoning
  for individualized treatment planning.
\end{abstract}

\section{Introduction}

Decision-makers are faced with the challenge of estimating what is likely to
happen when they take an action. One use of such an estimate is to evaluate
\emph{risk}; e.g. is this patient likely to die if I do not intervene? Another
use is to perform ``what~if?'' reasoning by comparing outcomes under alternative
actions; e.g. would changing the color or text of an ad lead to more
click-throughs? Practitioners commonly use supervised learning algorithms to
help decision-makers answer such questions, but these decision-support tools are
unreliable, and can even be dangerous.

Consider, for instance, the finding discussed by \citet{caruana2015intelligible}
regarding risk of death among those who develop pneumonia. Their goal was to
build a model that predicts risk of death for a hospitalized individual with
pneumonia so that those at high-risk could be treated and those at low-risk
could be safely sent home. Their model counterintuitively learned that
asthmatics are less likely to die from pneumonia. They traced the result back to
an \emph{existing policy} that asthmatics with pneumonia should be directly
admitted to the intensive care unit (ICU), therefore receiving more aggressive
treatment. Had this model been deployed to assess risk, then asthmatics might
have received \emph{less} care, putting them at greater
risk. \citet{caruana2015intelligible} show how these counterintuitive
relationships can be problematic and ought to be addressed by ``repairing'' the
model. We note, however, that these issues stem from a deeper limitation: when
training data is affected by actions, supervised learning algorithms capture
relationships caused by action policies, and these relationships do not
generalize when the policy changes.

To build reliable models for decision support, we propose using learning
objectives that predict \emph{counterfactuals}, which are collections of random
variables $\{Y[a] : a \in \mathcal{C}\}$ used in the \emph{potential outcomes}
framework
\citep{neyman1923applications,neyman1990application,rubin1978bayesian}. Counterfactuals
model the outcome $Y$ after an action $a$ is taken from a set of choices
$\mathcal{C}$. Counterfactual predictions are broadly applicable to a number of
decision-support tasks. In medicine, for instance, when evaluating a patient's
risk of death $Y$ to determine whether they should be treated aggressively, we
want an estimate of how they will fare \emph{without} treatment. This can be
done by predicting the counterfactual $Y[\varnothing]$, where $\varnothing$
stands for ``do nothing''. In online marketing, to decide whether we should
display ad $a_1$ or $a_2$, we may want an estimate of click-through $Y$ under
each, which amounts to predicting $Y[a_1]$ and $Y[a_2]$.

\begin{figure*}[t]
  \centering
  \includegraphics[width=0.9\linewidth]{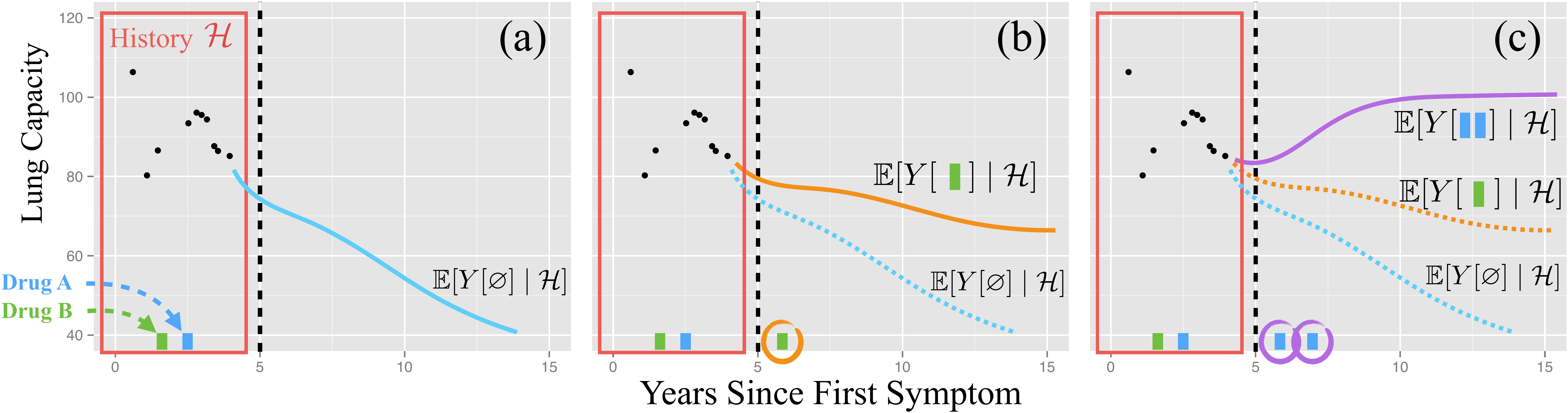}
  \caption{\footnotesize Best viewed in color. An illustration of the
    counterfactual GP applied to health care. The red box in (a) shows previous
    lung capacity measurements (black dots) and treatments (the history). Panels
    (a)-(c) show the type of predictions we would like to make. We use $Y[a]$ to
    represent the potential outcome under action $a$.}
  \label{fig:illustration}
\end{figure*}

To support decision-making in temporal settings, we develop the Counterfactual
Gaussian Process (CGP) to predict the counterfactual future progression of
continuous-time trajectories under sequences of future actions. The CGP can be
learned from and applied to time series data where actions are taken and
outcomes are measured at irregular time points; a generalization of discrete
time series. Figure \ref{fig:illustration} illustrates an application of the
CGP. We show an individual with a lung disease, and would like to predict her
future lung capacity (y-axis). Panel (a) shows the \emph{history} in the red
box, which includes previous lung capacity measurements (black dots) and
previous treatments (green and blue bars). The blue counterfactual trajectory
shows what might occur under \emph{no action}, which can be used to evaluate
this individual's risk. In panel (b), we show the counterfactual trajectory
under a single future green treatment. Panel (c) illustrates ``what~if?''
reasoning by overlaying counterfactual trajectories under two different action
sequences; in this case it seems that two future doses of the blue drug may lead
to a better outcome than a single dose of green.

\vspace{-8pt}
\paragraph{Contributions.}
Our key methodological contribution is the Counterfactual Gaussian process
(CGP), a model that predicts how a continuous-time trajectory will progress
under sequences of actions. We derive an adjusted maximum likelihood objective
that learns the CGP from \emph{observational traces}; irregularly sampled
sequences of actions and outcomes denoted using $\mathcal{D} = \{ \{ (y_{ij},
a_{ij}, t_{ij}) \}_{j=1}^{n_i} \}_{i=1}^m$, where $y_{ij} \in \R \cup
\set{\varnothing}$, $a_{ij} \in \mathcal{C} \cup \set{\varnothing}$, and $t_{ij}
\in [0, \tau]$.\footnote{$y_{ij}$ and $a_{ij}$ may be the null variable
  $\varnothing$ to allow for the possibility that an action is taken but no
  outcome is observed and vice versa. $[0, \tau]$ denotes a fixed period of time
  over which the trajectories are observed.} Our objective accounts for and
removes the effects of the policy used to choose actions in the observational
traces. We derive the objective by jointly modeling observed actions and
outcomes using a marked point process (MPP; see e.g.,
\citealt{daley2007introduction}), and show how it correctly learns the CGP under
a set of assumptions analagous to those required to learn counterfactual models
in other settings.

We demonstrate the CGP on two decision-support tasks. First, we show how the CGP
can make reliable risk predictions that do not depend on the action policy in
the training data. On the other hand, we show that predictions made by models
trained using classical supervised learning objectives are sensitive to the
policies.
In our second experiment, we use data from a real intensive care unit (ICU) to
learn the CGP, and qualitatively demonstrate how the CGP can be used to compare
counterfactuals and answer ``what~if?'' questions, which could offer medical
decision-makers a powerful new tool for individualized treatment planning.


\subsection{Related Work}

Decision support is a rich field; because our main methodological contribution
is a counterfactual model for time series data, we limit the scope of our
discussion of related work to this area.

\vspace{-8pt}
\paragraph{Causal inference.}
Counterfactual models stem from causal inference. In that literature, the
difference between the counterfactual outcomes if an action had been taken and
if it had not been taken is defined as the \emph{causal effect} of the action
(see e.g., \citealt{pearl2009causality} or \citealt{morgan2014counterfactuals}).
\emph{Potential outcomes} are commonly used to formalize counterfactuals and
obtain causal effect estimates
\citep{neyman1923applications,neyman1990application,rubin1978bayesian}. Potential
outcomes are often applied to cross-sectional data; see, for instance, the
examples in \citealt{morgan2014counterfactuals}. Recent examples from the
machine learning literature are \citet{bottou2013counterfactual} and
\citet{johansson2016learning}.

\vspace{-8pt}
\paragraph{Potential outcomes in discrete time.}
Potential outcomes have also been used to estimate the causal effect of a
sequence of actions in discrete time on a final outcome (e.g.
\citealt{robins1986new,robins2009estimation,taubman2009intervening}). The key
challenge in the sequential setting is to account for feedback between
intermediate outcomes that determine future treatment. Conversely,
\citet{brodersen2015inferring} estimate the effect that a \emph{single discrete
  intervention} has on a \emph{discrete} time series. Recent work on optimal
dynamic treatment regimes uses the sequential potential outcomes framework
proposed by \citet{robins1986new} to learn lists of discrete-time treatment
rules that optimize a scalar outcome. Algorithms for learning these rules often
use action-value functions (Q-learning; e.g., \citealt{nahum2012q}).
Alternatively, A-learning is a semiparametric approach that directly learns the
relative difference in value between alternative actions
\citep{murphy2003optimal}.

\vspace{-8pt}
\paragraph{Potential outcomes in continuous time.}
Others have extended the potential outcomes framework in \citet{robins1986new}
to learn causal effects of actions taken in continuous-time on a single final
outcome using observational data.  \citet{lok2008statistical} proposes an
estimator based on structural nested models \citep{robins1992estimation} that
learns the instantaneous effect of administering a single type of
treatment. \citet{arjas2004causal} develop an alternative framework for causal
inference using Bayesian posterior predictive distributions to estimate the
effects of actions in continuous time on a final outcome. Both
\citet{lok2008statistical} and \citet{arjas2004causal} use marked point
processes to formalize assumptions that make it possible to learn causal effects
from continuous-time observational data. We build on these ideas to learn causal
effects of actions on continuous-time \emph{trajectories} instead of a single
outcome. There has also been recent work on building expressive models of
treatment effects in continuous time. \citet{xu2016mlhc} propose a Bayesian
nonparametric approach to estimating individual-specific treatment effects of
discrete but irregularly spaced actions, and \citet{soleimani2017treatment}
model the effects of continuous-time, continuous-valued actions. Causal effects
in continuous-time have also been studied using differential
equations. \citet{mooij2013from} formalize an analog of Pearl's ``do'' operation
for deterministic ordinary differential equations. \citet{sokol2014causal} make
similar contributions for stochastic differential equations by studying limits
of discrete-time non-parametric structural equation models
\citep{pearl2009causality}.  \citet{cunningham2012gaussian} introduce the Causal
Gaussian Process, but their use of the term ``causal'' is different from ours,
and refers to a constraint that holds for sample paths of the GP.

\vspace{-8pt}
\paragraph{Reinforcement learning.}
Reinforcement learning (RL) algorithms learn from data where actions and
observations are interleaved in discrete time (see e.g.,
\citealt{sutton1998reinforcement}). In RL, however, the focus is on learning a
\emph{policy} (a map from states to actions) that optimizes the expected reward,
rather than a model that predicts the effects of the agent's actions on future
observations. In model-based RL, a model of an action's effect on the subsequent
state is produced as a by-product either offline before optimizing the policy
(e.g., \citealt{ng2006autonomous}) or incrementally as the agent interacts with
its environment. In most RL problems, however, learning algorithms rely on
active experimentation to collect samples. This is not always possible; for
example, in healthcare we cannot actively experiment on patients, and so we must
rely on retrospective observational data. In RL, a related problem known as
off-policy evaluation also uses retrospective observational data (see e.g.,
\citealt{dudik2011doubly,swaminathan2015counterfactual,jiang2016doubly,puaduraru2012empirical,doroudi2017importance}).
The goal is to use state-action-reward sequences generated by an agent operating
under an unknown policy to estimate the expected reward of a target
policy. Off-policy algorithms typically use action-value function approximation,
importance reweighting, or doubly robust combinations of the two to estimate the
expected reward.

\vspace{-7pt}
\section{Counterfactual Models from Observational Traces}
\label{sec:methods}
\vspace{-7pt}

Counterfactual GPs build on ideas from potential outcomes
\citep{neyman1923applications,neyman1990application,rubin1978bayesian}, Gaussian
processes \citep{rasmussen2006gaussian}, and marked point processes
\citep{daley2007introduction}. In the interest of space, we review potential
outcomes and marked point processes, but refer the reader to
\citet{rasmussen2006gaussian} for background on GPs.

\vspace{-8pt}
\paragraph{Background: Potential Outcomes.}
To formalize counterfactuals, we adopt the potential outcomes framework
\citep{neyman1923applications,neyman1990application,rubin1978bayesian}, which
uses a collection of random variables $\{ Y[a] : a \in \mathcal{C} \}$ to model
the outcome after each action $a$ from a set of choices $\mathcal{C}$.  To make
counterfactual predictions, we must learn the distribution $P(Y[a] \given X)$
for each action $a \in \mathcal{C}$ given features $X$. If we can freely
experiment by repeatedly taking actions and recording the effects, then it is
straightforward to fit a predictive model. Conducting experiments, however, may
not be possible. Alternatively, we can use observational data, where we have
example actions $A$, outcomes $Y$, and features $X$, but do not know how actions
were chosen. Note the difference between the action $a$ and the random variable
$A$ that models the \emph{observed actions} in our data; the notation $Y[a]$
serves to distinguish between the observed distribution $P(Y \given A, X)$ and
the target distribution $P(Y[a] \given X)$.

In general, we can only use observational data to estimate $P(Y \given A,
X)$. Under two assumptions, however, we can show that this conditional
distribution is equivalent to the counterfactual model $P(Y[a] \given X)$. The
first is known as the Consistency Assumption.
\begin{assumption}[Consistency]
  \label{as:consistency}
  Let $Y$ be the observed outcome, $A \in \mathcal{C}$ be the observed action,
  and $Y[a]$ be the potential outcome for action $a \in \mathcal{C}$, then: $
  (\, Y \triangleq Y[a] \,) \given A = a. $
\end{assumption}
\vspace{-5pt} Under consistency, we have that $P(Y \given A = a) = P(Y[a] \given
A = a)$. Now, the potential outcome $Y[a]$ may depend on the action $A$, so in
general $P(Y[a] \given A = a) \neq P(Y[a])$. The next assumption posits that the
features $X$ include all possible \emph{confounders}
\citep{morgan2014counterfactuals}, which are sufficient to d-separate $Y[a]$ and
$A$.
\begin{assumption}[No Unmeasured Confounders (NUC)]
  \label{as:nuc}
  Let $Y$ be the observed outcome, $A \in \mathcal{C}$ be the observed action,
  $X$ be a vector containing all potential confounders, and $Y[a]$ be the
  potential outcome under action $a \in \mathcal{C}$, then: $(\, Y[a] \ind A
  \,) \given X.$
\end{assumption}
\vspace{-5pt} Under Assumptions \ref{as:consistency} and \ref{as:nuc}, $P(Y
\given A, X) = P(Y[a] \given X)$. An extension of Assumption \ref{as:nuc}
introduced by \citet{robins1997causal} known as \emph{sequential NUC} allows us
to estimate the effect of a sequence of actions in discrete time on a single
outcome. In continuous-time settings, where both the type and \emph{timing} of
actions may be statistically dependent on the potential outcomes, Assumption
\ref{as:nuc} (and sequential NUC) cannot be applied as-is. We will describe an
alternative that serves a similar role for CGPs.

\vspace{-8pt}
\paragraph{Background: Marked Point Processes.}
Point processes are distributions over sequences of timestamps
$\set{T_i}_{i=1}^N$, which we call points, and a marked point process (MPP) is a
point process where each point is annotated with an additional random variable
$X_i$, called its mark. For example, a point $T$ might represent the arrival
time of a customer, and $X$ the amount that she spent at the store. We emphasize
that both the annotated points $(T_i, X_i)$ and the number of points $N$ are
random variables.

A point process can be characterized as a counting process $\{N_t : t \geq 0\}$
that counts the number of points that occured up to and including time $t$: $N_t
= \sum_{i=1}^N \1{T_i \leq t}$. By definition, this processes can only take
integer values, and $N_t \geq N_s$ if $t \geq s$. In addition, it is commonly
assumed that $N_0 = 0$ and that $\Delta N_t = \lim_{\delta \to 0^+} N_{t} - N_{t
  - \delta} \in \set{0, 1}$. We can parameterize a point process using a
probabilistic model of $\Delta N_t$ given the history of the process
$\mathcal{H}_{t^-}$ up to but not including time $t$ (we use $t^-$ to denote the
left limit of $t$). Using the Doob-Meyer decomposition
\citep{daley2007introduction}, we can write $\Delta N_t = \Delta M_t + \Delta
\Lambda_t$, where $M_t$ is a martingale, $\Lambda_t$ is a cumulative intensity
function, and
\begin{align*}
  P(\Delta N_t = 1 \given \mathcal{H}_{t^-})
  = \EE{ \Delta N_t \given \mathcal{H}_{t^-} }
  = \EE{ \Delta M_t \given \mathcal{H}_{t^-} } + \Delta \Lambda_t(\mathcal{H}_{t^-})
  = 0 + \Delta \Lambda_t(\mathcal{H}_{t^-}),
\end{align*}
which shows that we can parameterize the point process using the conditional
intensity function $\lambda^*(t) \d{t} \triangleq \Delta \Lambda_t
(\mathcal{H}_{t^-})$. The star superscript on the intensity function serves as a
reminder that it depends on the history $\mathcal{H}_{t^-}$. For example, in
non-homogeneous Poisson processes $\lambda^*(t)$ is a function of time that does
not depend on the history. On the other hand, a Hawkes process is an example of
a point process where $\lambda^*(t)$ \emph{does} depend on the history
\citep{hawkes1971spectra}. MPPs are defined by an intensity that is a function
of both the time $t$ and the mark $x$: $\lambda^*(t, x) = \lambda^*(t) p^*(x
\given t)$. We have written the joint intensity in a factored form, where
$\lambda^*(t)$ is the intensity of \emph{any} point occuring (that is, the mark
is unspecified), and $p^*(x \given t)$ is the pdf of the observed mark given the
point's time. For an MPP, the history $\mathcal{H}_t$ contains each prior
point's time and mark.

\subsection{Counterfactual Gaussian Processes}

Let $\set{Y_t : t \in [0, \tau]}$ denote a continuous-time stochastic process,
where $Y_t \in \R$, and $[0, \tau]$ defines the interval over which the process
is defined. We will assume that the process is observed at a discrete set of
irregular and random times $\set{(y_j, t_j)}_{j=1}^n$. We use $\mathcal{C}$ to
denote the set of possible \emph{action types}, $a \in \mathcal{C}$ to denote
the elements of the set, and define an action to be a 2-tuple $(a, t)$
specifying an action type $a \in \mathcal{C}$ and a time $t \in [0, \tau]$ at
which it is taken. To refer to multiple actions, we use $\vect{a} = [(a_1, t_1),
\ldots, (a_n, t_n)]$. Finally, we define the history $\mathcal{H}_t$ at a time
$t \in [0, \tau]$ to be a list of all previous observations of the process and
all previous actions. Our goal is to model the counterfactual:
\begin{align}
  \label{eq:target-counterfactual}
  P( \set{ Y_s[\vect{a}] : s > t } \given \mathcal{H}_t ),
  \text{   where $\vect{a} = \set{(a_j, t_j) : t_j > t}_{j=1}^m$.}
\end{align}
To learn the counterfactual model, we will use \emph{traces} ${\mathcal{D}
  \triangleq \{ \vect{h}_i = \set{(t_{ij}, y_{ij}, a_{ij})}_{j=1}^{n_i}
  \}_{i=1}^m}$, where $y_{ij} \in \R \cup \set{\varnothing}$, $a_{ij} \in
\mathcal{C} \cup \set{\varnothing}$, and $t_{ij} \in [0, \tau]$. Our approach is
to model $\mathcal{D}$ using a marked point process (MPP), which we learn using
the traces. Using Assumption \ref{as:consistency} and two additional assumptions
defined below, the estimated MPP recovers the counterfactual model in Equation
\ref{eq:target-counterfactual}.

We define the MPP mark space as the Cartesian product of the outcome space $\R$
and the set of action types $\mathcal{C}$. To allow either the outcome or the
action (but not both) to be the null variable $\varnothing$, we introduce binary
random variables $z_y \in \set{0, 1}$ and $z_a \in \set{0, 1}$ to indicate when
the outcome $y$ and action $a$ are not $\varnothing$. Formally, the mark space
is $\mathcal{X} = (\R \cup \set{\varnothing}) \times (\mathcal{C} \cup
\set{\varnothing}) \times \set{0, 1} \times \set{0, 1}$. We can then write the
MPP intensity as
\begin{align}
  \label{eq:mpp-intensity}
  \lambda^*(t, y, a, z_y, z_a) =
  \underbrace{\lambda^*(t) p^*(z_y, z_a \given t)}_{\text{[A] Event model}}
  \underbrace{p^*(y \given t, z_y)}_{\text{[B] Outcome model (GP)}}
  \underbrace{p^*(a \given y, t, z_a)}_{\text{[C] Action model}},
\end{align}
where we have again used the $*$ superscript as a reminder that the hazard
function and densities above are implicitly conditioned on the history
$\mathcal{H}_{t^-}$. The parameterization of the event and action models can be
chosen to reflect domain knowledge about how the timing of events and choice of
action depend on the history. The outcome model is parameterized using a GP (or
any elaboration such as a hierarchical GP or mixture of GPs), and can be treated
as a standard regression model that predicts how the future trajectory will
progress given the previous actions and outcome observations.

\vspace{-8pt}
\paragraph{Learning.}
To learn the CGP, we maximize the likelihood of observational traces over a
fixed interval $[0, \tau]$. Let $\vect{\theta}$ denote the model parameters,
then the likelihood for a single trace is
\begin{align}
  \label{eq:cgp-objective}
  \ell(\vect{\theta}) =
  \sum_{j=1}^n \log p^*_\theta(y_j \given t_j, z_{yj}) +
  \sum_{j=1}^n \log \lambda^*_\theta(t_j) p^*_\theta(a_j, z_{yj}, z_{aj} \given t_j, y_j) -
  \int_0^\tau \lambda^*_\theta(s) \d{s}.
\end{align}
We assume that traces are independent, and so can learn from multiple traces by
maximizing the sum of the individual-trace log likelihoods with respect to
$\vect{\theta}$. We refer to Equation \ref{eq:cgp-objective} as the adjusted
maximum likelihood objective. We see that the first term fits the GP to the
outcome data, and the second term acts as an adjustment to account for
dependencies between future outcomes and the timing and types of actions that
were observed in the training data.

\vspace{-8pt}
\paragraph{Connection to target counterfactual.}
By maximizing Equation \ref{eq:cgp-objective}, we obtain a statistical model of
the observational traces $\mathcal{D}$. In general, the statistical model may
not recover the target counterfactual model (Equation
\ref{eq:target-counterfactual}). To connect the CGP to Equation
\ref{eq:target-counterfactual}, we describe two additional assumptions. The
first assumption is an alternative to Assumption \ref{as:nuc}.
\begin{assumption}[Continuous-Time NUC]
  \label{as:ct-nuc}
  For all times $t$ and all histories $\mathcal{H}_{t^-}$, the densities
  $\lambda^*(t)$, $p^*(z_y, z_a \given t)$, and $p^*(a \given y, t, z_a)$ do not
  depend on $Y_s[\vect{a}]$ for all times $s > t$ and all actions $\vect{a}$.
\end{assumption}
\vspace{-5pt} The key implication of this assumption is that the policy used to
choose actions in the observational data did not depend on any unobserved
information that is predictive of the future potential outcomes.
\begin{assumption}[Non-Informative Measurement Times]
  \label{as:non-informative-measurement-times}
  For all times $t$ and any history $\mathcal{H}_{t^-}$, the following holds:
  $p^*(y \given t, z_y = 1) \d{y} = P( Y_t \in \d{y} \given \mathcal{H}_{t^-}).$
\end{assumption}
\vspace{-5pt} Under Assumptions \ref{as:consistency}, \ref{as:ct-nuc}, and
\ref{as:non-informative-measurement-times}, we can show that
Equation~\ref{eq:target-counterfactual} is equivalent to the GP used to model
$p^*(y \given t, z_y = 1)$. In the interest of space, the argument for this
equivalence is in Section \ref{sec:why-we-can-predict-potential-outcomes} of the
supplement. Note that these assumptions are not statistically testable (see
e.g., \citealt{pearl2009causality}).

\vspace{-7pt}
\section{Experiments}
\vspace{-7pt}

We demonstrate the CGP on two decision-support tasks. First, we show that the
CGP can make reliable risk predictions that are insensitive to the action policy
in the training data. Classical supervised learning algorithms, however, are
dependent on the action policy and this can make them unreliable
decision-support tools. Second, we show how the CGP can be used to compare
counterfactuals and ask ``what~if?'' questions for individualized treatment
planning by learning the effects of dialysis on creatinine levels using real
data from an intensive care unit (ICU).

\vspace{-5pt}
\subsection{Reliable Risk Prediction with CGPs}
\label{sec:reliable}
\vspace{-5pt}

\begin{table}
  \centering
  \begin{tabular}{r | r  r  | r  r  | r  r }
    \multicolumn{1}{c}{} & \multicolumn{2}{|c}{Regime $A$} & \multicolumn{2}{|c}{Regime $B$} & \multicolumn{2}{|c}{Regime $C$} \\
    & Baseline GP & CGP & Baseline GP & CGP & Baseline GP & CGP \\
    \midrule
    Risk Score $\Delta$ from $A$ & 0.000 & 0.000 & 0.083 & 0.001 & 0.162 & 0.128 \\
    Kendall's $\tau$ from $A$ & 1.000 & 1.000 & 0.857 & 0.998 & 0.640 & 0.562 \\
    AUC & 0.853 & 0.872 & 0.832 & 0.872 & 0.806 & 0.829 \\
  \end{tabular}
  \vspace{6pt}
  \caption{Results measuring reliability for simulated data experiments. See
    Section \ref{sec:reliable} for details.}
  \label{tab:simulation-results}
\end{table}

We first show how the CGP can be used for reliable risk prediction, where the
objective is to predict the likelihood of an adverse event so that we can
intervene to prevent it from happening. In this section, we use simulated data
so that we can evaluate using the true risk on test data. For concreteness, we
frame our experiment within a healthcare setting, but the ideas can be more
broadly applied. Suppose that a clinician records a real-valued measurement over
time that reflects an individual's health, which we call a \emph{severity
  marker}. We consider the individual to \emph{not} be at risk if the severity
marker is unlikely to fall below a particular threshold in the future without
intervention. As discussed by \citet{caruana2015intelligible}, modeling this
notion of risk can help doctors decide when an individual can safely be sent
home without aggressive treatment.


We simulate the value of a severity marker recorded over a period of 24 hours in
the hospital; high values indicate that the patient is healthy. A natural
approach to predicting risk at time $t$ is to model the conditional distribution
of the severity marker's future trajectory given the history up until time $t$;
i.e. $P(\set{Y_s : s > t} \given \mathcal{H}_t)$. We use this as our
baseline. As an alternative, we use the CGP to explicitly model the
counterfactual ``What if we do not treat this patient?'';
i.e. $P(\set{Y_s[\varnothing] : s > t} \given \mathcal{H}_t)$. For all
experiments, we consider a single decision time $t = 12\text{hrs}$. To quantify
risk, we use the negative of each model's predicted value at the end of 24
hours, normalized to lie in $[0, 1]$.

\vspace{-8pt}
\paragraph{Data.}
We simulate training and test data from three regimes. In regimes $A$ and $B$,
we simulate severity marker trajectories that are treated by policies $\pi_A$
and $\pi_B$ respectively, which are both unknown to the baseline model and CGP
at train time.  Both $\pi_A$ and $\pi_B$ are designed to satisfy Assumptions
\ref{as:consistency}, \ref{as:ct-nuc}, and
\ref{as:non-informative-measurement-times}. In regime $C$, we use a policy that
\emph{does not} satisfy these assumptions. This regime will demonstrate the
importance of verifying whether the assumptions hold when applying the CGP. We
train both the baseline model and CGP on data simulated from all three regimes.
We test all models on a common set of trajectories treated up until $t =
12\text{hrs}$ with policy $\pi_A$ and report how risk predictions vary as a
function of action policy in the training data.

\vspace{-8pt}
\paragraph{Simulator.}
For each patient, we randomly sample outcome measurement times from a
homogeneous Poisson process with with constant intensity $\lambda$ over the 24
hour period. Given the measurement times, outcomes are sampled from a mixture of
three GPs. The covariance function is shared between all classes, and is defined
using a Mat\'{e}rn $3/2$ kernel (variance $0.2^2$, lengthscale $8.0$) and
independent Gaussian noise (scale $0.1$) added to each observation. Each class
has a distinct mean function parameterized using a 5-dimensional, order-3
B-spline. The first class has a declining mean trajectory, the second has a
trajectory that declines then stabilizes, and the third has a stable
trajectory.\footnote{The exact B-spline coefficients can be found in the
  simulation code included in the supplement.} All classes are equally likely
\emph{a priori}. At each measurement time, the treatment policy $\pi$ determines
a probability $p$ of treatment administration (we use only a single treatment
type). The treatments increase the severity marker by a constant amount for 2
hours. If two or more actions occur within 2 hours of one another, the effects
do not add up (i.e. it is as though only one treatment is active). Additional
details about the simulator and policies can be found in the supplement.

\vspace{-8pt}
\paragraph{Model.}
For both the baseline GP and CGP, we use a mixture of three GPs (as was used to
simulate the data). We assume that the mean function coefficients, the
covariance parameters, and the treatment effect size are unknown and must be
learned. We emphasize that both the baseline GP and CGP have identical forms,
but are trained using different objectives; the baseline marginalizes over
future actions, inducing a dependence on the treatment policy in the training
data, while the CGP explicitly controls for them while learning. For both the
baseline model and CGP, we analytically sum over the mixture component
likelihoods to obtain a closed form expression for the likelihood, which we
optimize using BFGS \citep{nocedal2006numerical}.\footnote{Additional details
  can be found in the supplement.} Predictions for both models are made using
the posterior predictive mean given data and interventions up until 12 hours.

\vspace{-8pt}
\paragraph{Results.}
We find that the baseline GP's risk scores fluctuate across regimes $A$, $B$,
and $C$. The CGP is stable across regimes $A$ and $B$, but unstable in regime
$C$, where our assumptions are violated. In Table \ref{tab:simulation-results},
the first row shows the average difference in risk scores (which take values in
$[0, 1]$) produced by the models trained in each regime and produced by the
models trained in regime $A$. In row 1, column $B$ we see that the baseline GP's
risk scores differ for the same person on average by around eight points
($\Delta = 0.083$). From the perspective of a decision-maker, this behavior
could make the system appear less reliable. Intuitively, the risk for a given
patient should not depend on the policy used to determine treatments in
retrospective data. On the other hand, the CGP's scores change very little when
trained on different regimes ($\Delta = 0.001$), as long as Assumptions
\ref{as:consistency}, \ref{as:ct-nuc}, and
\ref{as:non-informative-measurement-times} are satisfied.

A cynical reader might ask: even if the risk scores are unstable, perhaps it has
no consequences on the downstream decision-making task? In the second row of
Table \ref{tab:simulation-results}, we report Kendall's $\tau$ computed between
each regime and regime $A$ using the risk scores to rank the patient's in the
test data according to severity (i.e. scores closer to $1$ are more severe). In
the third row, we report the AUC for both models trained in each regime on the
common test set. We label a patient as ``at risk'' if the last marker value in
the untreated trajectory is below zero, and ``not at risk'' otherwise. In row 2,
column $B$ we see that the CGP has a high rank correlation ($\tau = 0.998$)
between the two regimes where the policies satisfy our key assumptions.  The
baseline GP model trained on regime $B$, however, has a lower rank correlation
of $\tau = 0.857$ with the risk scores produced by the same model trained on
regime $A$. Similarly, in row three, columns $A$ and $B$, we see that the CGP's
AUC is unchanged ($\text{AUC} = 0.872$). The baseline GP, however, is unstable
and creates a risk score with poorer discrimination in regime $B$ ($\text{AUC} =
0.832$) than in regime $A$ ($\text{AUC} = 0.853$).  Although we illustrate
stability of the CGP compared to the baseline GP using two regimes, this
property is not specific to the particular choice of policies used in regimes
$A$ and $B$; the issue persists as we generate different training data by
varying the distribution over the action choices.

Finally, the results in column $C$ highlight the importance of Assumptions
\ref{as:consistency}, \ref{as:ct-nuc}, and
\ref{as:non-informative-measurement-times}. The policy $\pi_C$ \emph{does not}
satisfy these assumptions, and we see that the risk scores for the CGP are
different when fit in regime $C$ than when fit in regime $A$ ($\Delta =
0.128$). Similarly, in row 2 the CGP's rank correlation degrades ($\tau =
0.562$), and in row 3 the AUC decreases to $0.829$. Note that the baseline GP
continues to be unstable when fit in regime $C$.

\vspace{-8pt}
\paragraph{Conclusions.}
These results have important implications for the practice of building
predictive models for decision support. Classical supervised learning algorithms
can be unreliable due to an implicit dependence on the action policy in the
training data, which is usually different from the assumed action policy at test
time (e.g. what will happen if we do not treat?).
Note that this issue is not resolved by training only on individuals who are not
treated because selection bias creates a mismatch between our train and test
distributions. From a broader perspective, supervised learning can be unreliable
because it captures features of the training distribution that may change
(e.g. relationships caused by the action policy). Although we have used a
counterfactual model to account for and remove these unstable relationships,
there may be other approaches that achieve the same effect (e.g.,
\citealt{dyagilev2016learning}). Recent related work by \citet{gong2016domain}
on covariate shift aims to learn only the components of the source distribution
that will generalize to the target distribution. As predictive models are
becoming more widely used in domains like healthcare where safety is critical
(e.g. \citealt{\riskpapers}), the framework proposed here is increasingly
pertinent.

\subsection{``What~if?'' Reasoning for Individualized Treatment Planning}

\begin{figure}[t]
  \centering
  \includegraphics[width=1.0\linewidth]{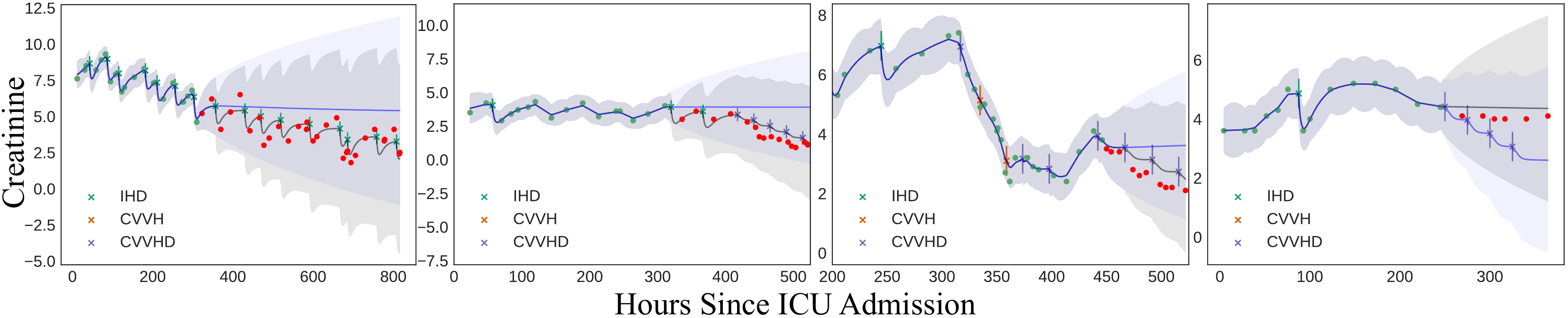}
  \caption{\small Example factual (grey) and counterfactual (blue) predictions
    on real ICU data using the CGP.}
  \label{fig:icu-results}
\end{figure}


To demonstrate how the CGP can be used for individualized treatment planning, we
extract observational creatinine traces from the publicly available MIMIC-II
database \citep{saeed2011multiparameter}. Creatinine is a compound produced as a
by-product of the chemical reaction in the body that breaks down creatine to
fuel muscles. Healthy kidneys normally filter creatinine out of the body, which
can otherwise be toxic in large concentrations. During kidney failure, however,
creatinine levels rise and the compound must be extracted using a medical
procedure called dialysis.

We extract patients in the database who tested positive for abnormal creatinine
levels, which is a sign of kidney failure. We also extract the times at which
three different types of dialysis were given to each individual: intermittent
hemodialysis (IHD), continuous veno-venous hemofiltration (CVVH), and continuous
veno-venous hemodialysis (CVVHD). The data set includes a total of 428
individuals, with an average of 34 ($\pm 12$) creatinine observations each. We
shuffle the data and use 300 traces for training, 50 for validation and model
selection, and 78 for testing.

\vspace{-8pt}
\paragraph{Model.}
We parameterize the outcome model of the CGP using a mixture of GPs. We always
condition on the initial creatinine measurement and model the deviation from
that initial value. The mean for each class is zero (i.e. we assume there is no
deviation from the initial value on average). We parameterize the covariance
function using the sum of two non-stationary kernel functions. Let $\phi : t \to
[1, t, t^2]^\T \in \R^3$ denote the quadratic polynomial basis, then the first
kernel is $k_1(t_1, t_2) = \phi^\T(t_1) \matr{\Sigma} \phi(t_2)$, where
$\matr{\Sigma} \in \R^{3 \times 3}$ is a positive-definite symmetric matrix
parameterizing the kernel. The second kernel is the covariance function of the
integrated Ornstein-Uhlenbeck (IOU) process (see e.g.,
\citealt{taylor1994stochastic}), which is parameterized by two scalars $\alpha$
and $\nu$ and defined as
\begin{align}
  \nonumber
  \textstyle
  k_{\text{IOU}}(t_1, t_2) = \frac{\nu^2}{2 \alpha^3}
  \left(
  2 \alpha \text{min}(t_1, t_2) + e^{-\alpha t_1} + e^{-\alpha t_2} - 1 - e^{-\alpha | t_1 - t_2 |}
  \right).
\end{align}
The IOU covariance corresponds to the random trajectory of a particle whose
velocity drifts according to an OU process. We assume that each creatinine
measurement is observed with independent Gaussian noise with scale $\sigma$.
Each class in the mixture has a unique set of covariance parameters. To model
the treatment effects in the outcome model, we define a short-term function and
long-term response function. If an action is taken at time $t_0$, the outcome
$\delta = t - t_0$ hours later will be additively affected by the response
function $g(\delta ; h_1, a, b, h_2, r) = g_s(\delta ; h_1, a, b) +
g_\ell(\delta ; h_2, r)$, where $h_1, h_2 \in \R$ and $a, b, r \in \R^+$. The
short-term and long-term response functions are defined as $g_s(\delta; h_1, a,
b) = \frac{h_1 a}{a - b} \left( e^{-b \cdot t} - e^{-a \cdot t} \right)$, and
$g_\ell(\delta: h_2, r) = h_2 \cdot \left( 1.0 - e^{-r \cdot t} \right)$. The
two response functions are included in the mean function of the GP, and each
class in the mixture has a unique set of response function parameters. We assume
that Assumptions \ref{as:consistency}, \ref{as:ct-nuc}, and
\ref{as:non-informative-measurement-times} hold, and that the event and action
models have separate parameters, so can remain unspecified when estimating the
outcome model. We fit the CGP outcome model using Equation
\ref{eq:cgp-objective}, and select the number of classes in the mixture using
fit on the validation data (we choose three components).

\vspace{-8pt}
\paragraph{Results.}
Figure \ref{fig:icu-results} demonstrates how the CGP can be used to do
``what~if?'' reasoning for treatment planning. Each panel in the figure shows
data for an individual drawn from the test set. The green points show
measurements on which we condition to obtain a posterior distribution over
mixture class membership and the individual's latent trajectory under each
class. The red points are unobserved, future measurements. In grey, we show
predictions under the \emph{factual} sequence of actions extracted from the
MIMIC-II database. Treatment times are shown using vertical bars marked with an
``x'' (color indicates which type of treatment was given). In blue, we show the
CGP's \emph{counterfactual} predictions under an alternative sequence of
actions. The posterior predictive trajectory is shown for the MAP mixture class
(mean is shown by a solid grey/blue line, $95\%$ credible intervals are shaded).

We qualitatively discuss the CGP's counterfactual predictions, but cannot
quantitatively evaluate them without prospective experimental data from the ICU.
We can, however, measure fit on the factual data and compare to baselines to
evaluate our modeling decisions. Our CGP's outcome model allows for
heterogeneity in the covariance parameters and the response functions. We
compare this choice to two alternatives. The first is a mixture of three GPs
that \emph{does not} model treatment effects. The second is a single GP that
\emph{does} model treatment effects. Over a 24-hour horizon, the CGP's mean
absolute error (MAE) is 0.39 ($95\%$ CI: 0.38-0.40),\footnote{$95\%$ confidence
  intervals computed using the pivotal bootstrap are shown in parentheses}, and
for predictions between 24 and 48 hours in the future the MAE is 0.62 ($95\%$
CI: 0.60-0.64). The pairwise mean difference between the first baseline's
absolute errors and the CGP's is 0.07 (0.06, 0.08) for 24 hours, and 0.09 (0.08,
0.10) for 24-48 hours. The mean difference between the second baseline's
absolute errors and the CGP's is 0.04 (0.04, 0.05) for 24 hours and 0.03 (0.02,
0.04) for 24-48 hours. The improvements over the baselines suggest that modeling
treatments and heterogeneity with a mixture of GPs for the outcome model are
useful for this problem.

Figure \ref{fig:icu-results} shows factual and counterfactual predictions made
by the CGP. In the first (left-most) panel, the patient is factually
administered IHD about once a day, and is responsive to the treatment
(creatinine steadily improves). We query the CGP to estimate how the individual
\emph{would have} responded had the IHD treatment been stopped early. The model
reasonably predicts that we would have seen no further improvement in
creatinine. The second panel shows a similar case. In the third panel, an
individual with erratic creatinine levels receives CVVHD for the last 100 hours
and is responsive to the treatment. As before, the CGP counterfactually predicts
that she would not have improved had CVVHD not been given. Interestingly, panel
four shows the opposite situation: the individual did not receive treatment and
did not improve for the last 100 hours, but the CGP counterfactually predicts an
improvement in creatinine as in panel 3 under daily CVVHD.

\vspace{-8pt}
\section{Discussion}
\vspace{-9pt}

Classical supervised learning algorithms can lead to unreliable and, in some
cases, dangerous decision-support tools. As a safer alternative, this paper
advocates for using potential outcomes
\citep{neyman1923applications,neyman1990application,rubin1978bayesian} and
\emph{counterfactual learning objectives} (like the one in Equation
\ref{eq:cgp-objective}). We introduced the Counterfactual Gaussian Process (CGP)
as a decision-support tool for scenarios where outcomes are measured and actions
are taken at irregular, discrete points in continuous-time. The CGP builds on
previous ideas in continuous-time causal inference
(e.g. \citealt{robins1997causal,arjas2004causal,lok2008statistical}), but is
unique in that it can predict the full counterfactual \emph{trajectory} of a
time-dependent outcome. We designed an adjusted maximum likelihood algorithm for
learning the CGP from \emph{observational traces} by modeling them using a
marked point process (MPP), and described three structural assumptions that are
sufficient to show that the algorithm correctly recovers the CGP.

We empirically demonstrated the CGP on two decision-support tasks. First, we
showed that the CGP can be used to make reliable risk predictions that are
insensitive to the action policies used in the training data. This is critical
because an action policy can cause a predictive model fit using classical
supervised learning to capture relationships between the features and outcome
(risk) that lead to poor downstream decisions and that are difficult to
diagnose. In the second set of experiments, we showed how the CGP can be used to
compare counterfactuals and answer ``what~if?'' questions, which could offer
decision-makers a powerful new tool for individualized treatment planning. We
demonstrated this capability by learning the effects of dialysis on creatinine
trajectories using real ICU data and predicting counterfactual progressions
under alternative dialysis treatment plans.

These results suggest a number of new questions and directions for future work.
First, the validity of the CGP is conditioned upon a set of assumptions (this is
true for all counterfactual models). In general, these assumptions are not
testable. The reliability of approaches using counterfactual models therefore
critically depends on the plausibility of those assumptions in light of domain
knowledge. Formal procedures, such as sensitivity analyses (e.g.,
\citealt{robins2000sensitivity,scharfstein2014global}), that can identify when
causal assumptions conflict with a data set will help to make these methods more
easily applied in practice. In addition, there may be other sets of structural
assumptions beyond those presented that allow us to learn counterfactual GPs
from non-experimental data. For instance, the back door and front door criteria
are two separate sets of structural assumptions discussed by
\citet{pearl2009causality} in the context of estimating parameters of causal
Bayesian networks from observational data.

More broadly, this work has implications for recent pushes to introduce safety,
accountability, and transparency into machine learning systems. We have shown
that learning algorithms sensitive to certain factors in the training data (the
action policy, in this case) can make a system less reliable. In this paper, we
used the potential outcomes framework and counterfactuals to characterize and
account for such factors, but there may be other ways to do this that depend on
fewer or more realistic assumptions (e.g.,
\citealt{dyagilev2016learning}). Moreover, removing these nuisance factors is
complementary to other system design goals such as interpretability (e.g.,
\citealt{ribeiro2016should}).

\vspace{-8pt}
\subsubsection*{Acknowledgements}
\vspace{-8pt}

We thank the anonymous reviewers for their insightful feedback. This work was
supported by generous funding from DARPA YFA \#D17AP00014 and NSF SCH \#1418590.
PS was also supported by an NSF Graduate Research Fellowship. We thank Katie
Henry and Andong Zhan for help with the ICU data set. We also thank Miguel
Hern\'{a}n for pointing us to earlier work by James Robins on
treatment-confounder feedback.

\appendix
\section{Equivalence of MPP Outcome Model and Counterfactual Model}
\label{sec:why-we-can-predict-potential-outcomes}

At a given time $t$, we want to make predictions about the potential outcomes
that we will measure at a set of future query times $\vect{q} = [s_1, \ldots,
s_m]$ given a specified future sequence of actions $\vect{a}$. This can be
written formally as
\begin{align}
  P(\set{Y_s[\vect{a}] : s \in \vect{q}} \given \mathcal{H}_t)
\end{align}

Without loss of generality, we can use the chain rule to factor this joint
distribution over the potential outcomes. We choose a factorization in time
order; that is, a potential outcome is conditioned on all potential outcomes at
earlier times. We now describe a sequence of steps that we can apply to each
factor in the product.
\begin{align}
  \label{eq:factored-query}  
  P(\set{Y_s[\vect{a}] : s \in \vect{q}} \given \mathcal{H}_t)
  =
  \prod_{i=1}^m
  P(Y_{s_i}[\vect{a}] \given \set{Y_{s}[\vect{a}] : s \in \vect{q}, s < s_i}, \mathcal{H}_t).
\end{align}

Using Assumption \ref{as:ct-nuc}, we can introduce random variables for marked
points that have the same timing and actions as the proposed sequence of actions
without changing the probability. Recall our assumption that actions can only
affect future values of the outcome, so we only need to introduce marked points
for actions taken at earlier times. Formally, we introduce the set of marked
points for the potential outcome at each time $s_i$
\begin{align}
  \vect{A}_i = \set{(t', \varnothing, a, 0, 1) : (t', a) \in \vect{a}, t' < s_i}.
\end{align}
We can then write
\begin{align}
  P(Y_{s_i}[\vect{a}] \given \set{Y_{s}[\vect{a}] : s \in \vect{q}, s < s_i}, \mathcal{H}_t)
  = P(Y_{s_i}[\vect{a}] \given \vect{A}_i, \set{Y_{s}[\vect{a}] : s \in \vect{q}, s < s_i}, \mathcal{H}_t).
\end{align}
To show that $P(Y[a] \given A = a, X = x) = P(Y[a] \given X = x)$ in Section
\ref{sec:methods}, we use Assumption \ref{as:nuc} to remove the random variable
$A$ from the conditioning information without changing the probability
statement. We reverse that logic here by adding $\vect{A}_i$.

Now, under Assumption \ref{as:consistency}, after conditioning on $\vect{A}_i$,
we can replace the potential outcome $Y_{s_i}[\vect{a}]$ with $Y_{s_i}$. We
therefore have
\begin{align}
  P(Y_{s_i}[\vect{a}] \given \vect{A}_i, \set{Y_{s}[\vect{a}] : s \in \vect{q}, s < s_i}, \mathcal{H}_t)
  = P(Y_{s_i} \given \vect{A}_i, \set{Y_{s}[\vect{a}] : s \in \vect{q}, s < s_i}, \mathcal{H}_t).
\end{align}

Similarly, because the set of proposed actions affecting the outcome at time
$s_i$ contain all actions that affect the outcome at earlier times $s < s_i$, we
can invoke Assumption \ref{as:consistency} again and replace all potential
outcomes at earlier times with the value of the observed process at that time.
\begin{align}
  \nonumber
  P(Y_{s_i} \given \vect{A}_i, \set{Y_{s}[\vect{a}] : s \in \vect{q}, s < s_i}, \mathcal{H}_t)
  = P(Y_{s_i} \given \vect{A}_i, \set{Y_{s} : s \in \vect{q}, s < s_i}, \mathcal{H}_t).
\end{align}

Next, Assumption \ref{as:non-informative-measurement-times} posits that the
outcome model ${p^*(y \given t', z_y = 1)}$ is the density of ${P(Y_{t'} \given
  \mathcal{H}_t)}$, which implies that the mark $(t', y, \varnothing, 1, 0)$ is
equivalent to the event $(Y_{t'} \in \d{y})$. Therefore, for each $s_i$ define
\begin{align}
  \vect{O}_i = \set{(s, Y_s, \varnothing, 1, 0) : s \in \vect{q}, s < s_i}.
\end{align}
Using this definition, we can write
\begin{align}
  \nonumber
  P(Y_{s_i} \given \vect{A}_i, \set{Y_{s} : s \in \vect{q}, s < s_i}, \mathcal{H}_t)
  = (Y_{s_i} \given \vect{A}_i, \vect{O}_i, \mathcal{H}_t).
\end{align}

The set of information $(\vect{A}_i, \vect{O}_i, \mathcal{H}_t)$ is a valid
history of the marked point process $\mathcal{H}_{s_i}^-$ up to but not
including time $s_i$. We can therefore replace all information after the
conditioning bar in each factor of Equation \ref{eq:factored-query} with
$\mathcal{H}_{s_i^-}$.
\begin{align}
  P(Y_{s_i} \given \vect{A}_i, \vect{O}_i, \mathcal{H}_t)
  = P(Y_{s_i} \given \mathcal{H}^-_{s_i}).
\end{align}
Finally, by applying Assumption \ref{as:non-informative-measurement-times}
again, we have
\begin{align}
  P(Y_{s_i} \in \d{y} \given \mathcal{H}^-_{s_i})
  = p^*(y \given s_i, z_y = 1) \d{y}.
\end{align}
The potential outcome query can therefore be answered using the outcome model,
which we can estimate from data.

\section{Causal Bayesian Network}

\begin{figure*}[t]
  \centering
  \includegraphics[width=0.8\linewidth]{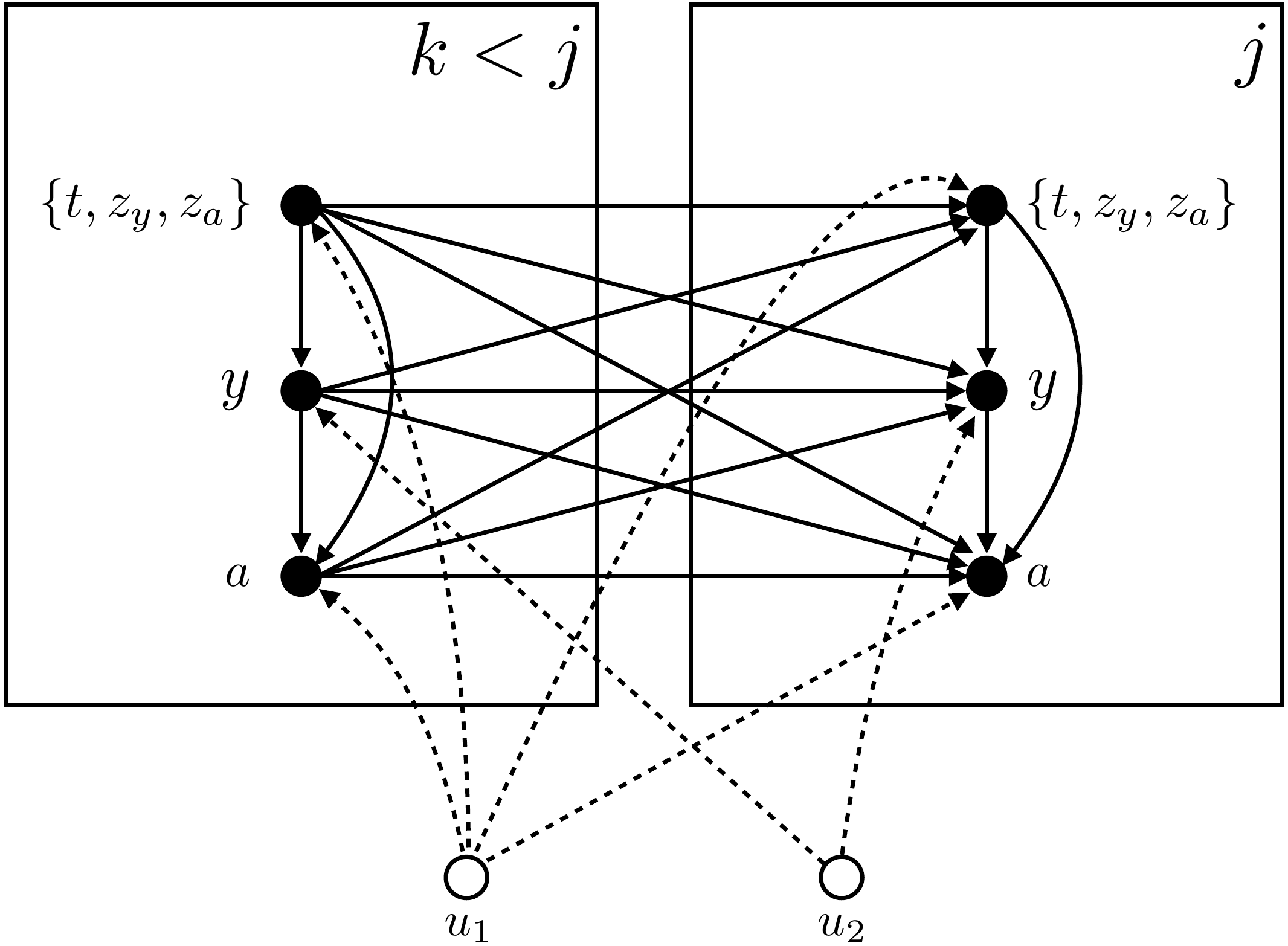}
  \caption{The causal Bayesian network for the counterfactual GP.}
  \label{fig:causal-dag}
\end{figure*}

We can also characterize our key assumptions using causal Bayesian networks
\citep{pearl2009causality}. Let $\{(t_j, z_{y,j}, z_{a,j}, y_j, a_j)\}_{j \geq
  1}$ be a countable sequence of tuples of variables (a marked point process can
be characterized as a countable sequence of points and marks). Recall that $t_j$
is an event time, $z_{y,j}$ is a binary random variable indicating whether an
outcome is measured, $z_{a,j}$ is a binary random variable indicating whether an
action is taken, $y_j \in \mathcal{R} \cup \{\varnothing\}$ is an outcome
measurement, and $a_j \in \mathcal{C} \cup \{\varnothing\}$ is an action (the
last two variables are $\varnothing$ when the respective indicator is $0$).

We define the directed acyclic graph $\mathcal{G}$ with nodes $\mathcal{V}
\triangleq \cup_{j \geq 1} \{t_j, z_{y,j}, z_{a,j}, y_j, a_j\}$ and edge set
$\mathcal{E}$ to be the causal Bayesian network for the counterfactual GP. For
any variables $v_1 \in \{t_j, z_{y,j}, z_{a,j}, y_j, a_j\}$ and $v_2 \in \{t_k,
z_{y,k}, z_{a,k}, y_k, a_k\}$, the edge $(v_1 \to v_k) \in \mathcal{E}$ if $j <
k$ or if $j = k$ and $v_1$ is a parent of $v_2$ in the right-most plate of
Figure \ref{fig:causal-dag}. We allow the variables $\{(t_j, z_{y,j}, z_{a,j},
a_j)\}_{j=1}^\infty$ to depend on a common unobserved parent $u_1$, and the
outcomes $\{y_j\}_{j=1}^\infty$ to depend on a common unobserved parent $u_2$.
The DAG in Figure \ref{fig:causal-dag} sketches the causal Bayesian network. For
any index $j$, we show the edges present between all variables at times $k < j$.

We now formulate our causal query, and show that it is identified using
observational traces sampled from the distribution implied by the causal
Bayesian network. For any time $t \in [0, \tau]$, our goal is to predict the
values of future outcomes under a hypothetical sequence of future actions given
the history up until time $t$. Define $\mathcal{H}_t = \cup_{j : t_j < t} \{t_j,
z_{y,j}, z_{a,j}, y_j, a_j\}$ to be the sequence of $n$ actions taken and
outcomes measured prior to time $t$, and define $\mathcal{F}_t$ to be a sequence
of $m$ tuples corresponding to future actions and measurements. The variables in
$\mathcal{H}_t \cup \mathcal{F}_t$ are connected using the edge set definition
described above. Let $\bm{t}$ denote the $m$ future time points, $\bm{z}_y$ the
future measurement indicators, $\bm{z}_a$ the future action indicators, $\bm{y}$
the future outcomes, and $\bm{a}$ the future actions. Our goal is to show that
the following query is identified:
\begin{align}
  p(\bm{y} \mid \text{do}(\bm{t}, \bm{z}_y, \bm{z}_a, \bm{a}), \mathcal{H}_t)
  = \prod_{j=1}^m
  p(
  y_j
  \mid
  \bm{\bar{y}}_{:j},
  \text{do}(\bm{t}, \bm{z}_y, \bm{z}_a, \bm{a}),
  \mathcal{H}_t
  ),
\end{align}
where $\bm{\bar{y}}_{:j}$ denotes the vector of future outcomes before the
$j^{\text{th}}$. We will also use $\bm{\bar{y}}_{j:}$ to denote all outcomes
measured after the $j^{\text{th}}$ (this notation will be used for the other
variables as well). First, consider any factor in the expression above. We
define the future and past intervened-on variables at time $t_j$ as
\begin{gather}
  \bm{f}_j \triangleq
  \{
  a_j,
  \bm{\bar{t}}_{j:},
  \bm{\bar{z}}_{y,j:},
  \bm{\bar{z}}_{a,j:},
  \bm{\bar{a}}_{j:}
  \} \\
  \bm{p}_j \triangleq
  \{
  \bm{\bar{t}}_{:j},
  \bm{\bar{z}}_{y,:j},
  \bm{\bar{z}}_{a,:j},
  \bm{\bar{a}}_{:j},
  t_j, z_{y,j}, z_{a,j}
  \}.             
\end{gather}
Using these shorthand definitions, we first prove the following equivalence
\begin{align}
  p(
  y_j
  \mid
  \bm{\bar{y}}_{:j},
  \text{do}(\bm{p}_j),
  \text{do}(\bm{f}_j),
  \mathcal{H}_t
  )
  =
  p(
  y_j
  \mid
  \bm{\bar{y}}_{:j},
  \text{do}(\bm{p}_j),
  \mathcal{H}_t
  ).
\end{align}
Intuitively, we are showing that actions taken after $y_j$ is measured do not
affect its value. To justify the equality, we use ``Rule 3'' from Pearl's
do-calculus (see Chapter 3 in \citealt{pearl2009causality}). We must show that
$y_j$ is d-separated from $\bm{f}_j$ in the mutilated DAG where all incoming
edges to nodes in $\bm{p}_j$ and $\bm{f}_j$ have been removed. To show
d-separation, let $v \in \bm{f}_j \setminus \{a_j\}$ be some future
intervened-on variable at time step $k > j$. Since all incoming edges have been
removed, all paths starting at $v$ must be outgoing. Outgoing edges for $v$ in
the original DAG either point to an outcome $y_\ell$ for $\ell \geq k$ or some
other intervened-on variable $v' \in \bm{f}_j \setminus \{a_j, v\}$. The latter
are removed in the mutilated graph, so the only edges outgoing from $v$ must
point to an outcome $y_\ell$ for $\ell \geq k$. This implies that all paths
starting at $v$ must begin with an edge $v \to y_\ell$ for some $\ell \geq k$.
Because $y_\ell$ is unobserved, the only unblocked paths must then follow an
outgoing edge (otherwise it would be a collider). All outgoing edges from
variables $y_\ell$ for $\ell \geq k$ can only point to outcomes $y_{\ell'}$ for
$\ell' > \ell$, which in turn must point to $y_{\ell''}$ for $\ell'' > \ell'$,
and so on. Therefore, any path starting from $v$ must pass through outcomes $y$
at strictly increasing times. Eventually, we will reach the final outcome, where
there are no outgoing edges, ending the path. We can conclude that no paths
starting at $v$ can reach $y_j$. A similar argument shows that no path starting
from $a_j$ can reach $y_j$.

Next, we use ``Rule 2'' from the do-calculus to prove that
\begin{align}
  p(
  y_j
  \mid
  \bm{\bar{y}}_{:j},
  \text{do}(\bm{p}_j),
  \mathcal{H}_t
  )
  =
  p(
  y_j
  \mid
  \bm{\bar{y}}_{:j},
  \bm{p}_j,
  \mathcal{H}_t
  ).
\end{align}
This requires showing that $y_j$ is d-separated from $\bm{p}_j$ in the mutilated
graph where all outgoing edges from $v \in \bm{p}_j$ have been removed. For any
$v \in \bm{p}_j$, there are two types of incoming edges. The first are edges
originating from observed direct parents of $v$, and the second is the edge
originating from the unobserved variable $u_1$. Any path from $v$ to $y_j$ must
start with one of these edge types, and therefore all that start with an edge to
an observed parent of $v$ will be blocked, and any unblocked path must start by
going through $u_1$. Now, $u_1$ has no parents and any path must then have a
second edge from $u_1$ to one of its children, which are all times $t_k$,
indicators $z_{y,k}$ or $z_{a,k}$, and actions $a_k$. We will analyze these
possibilities using two cases. First, the second edge could go from $u_1$ to a
time $t_k$ where $k \leq j$, indicator $z_{y,k}$ or $z_{a,k}$ where $k \leq j$,
or to an action $a_k$ where $k < j$. The only possible next step is to go
through an incoming edge where the origin is not $u_1$; all such edges will be
blocked, and so cannot reach $y_j$. In the second case, an edge could go from
$u_1$ to a time or indicator at step $k > j$, or an action at step $k \geq j$.
These variables are unobserved, and so the only valid next step is to follow an
outgoing edge. Subsequent steps must all also follow outgoing edges by the same
logic, and so the path can never return to $y_j$. We therefore can conclude that
there are no paths from $v \in \bm{p}_j$ to $y_j$ in the mutilated graph, so the
equality holds. Together, the two inequalities show
\begin{align}
  p(\bm{y} \mid \text{do}(\bm{t}, \bm{z}_y, \bm{z}_a, \bm{a}), \mathcal{H}_t)
  = \prod_{j=1}^m
  p(
  y_j
  \mid
  \bm{\bar{y}}_{:j},
  \bm{p}_j,
  \mathcal{H}_t
  ).
\end{align}
This shows that the structural dependencies encoded in the graph shown in Figure
\ref{fig:causal-dag} can be used in place of Assumption \ref{as:ct-nuc}. In
addition, we no longer need Assumption \ref{as:consistency} (consistency), which
highlights an interesting difference between the potential outcomes and causal
Bayesian network frameworks. In Pearl's causal DAGs, consistency is in fact a
theorem derived from the axioms of the framework, whereas it is assumed in the
potential outcomes framework. This is shown in Corollary 7.3.2 in
\citet{pearl2009causality}, which follows from the Composition axiom and the
definition of a ``null'' intervention. Intuitively, the fact that consistency is
a theorem in Pearl's framework reflects the assumption that the parent-child
relationships in the DAG are sufficiently stable, autonomous, or ``local''
\citep{pearl2009causality}. See Section 7.2.4 in \citet{pearl2009causality} for
further information. Finally, Assumption
\ref{as:non-informative-measurement-times} remains unchanged and simply allows
us to treat measured outcomes $y_j$ as unbiased samples of the process
$Y_{t_j}$.

\section{Simulation and Policy Details}

For each patient, we randomly sample outcome measurement times from a
homogeneous Poisson process with with constant intensity $\lambda$ over the 24
hour period. Given the measurement times, outcomes are sampled from a mixture of
three GPs. The covariance function is shared between all classes, and is defined
using a Mat\'{e}rn $3/2$ kernel (variance $0.2^2$, lengthscale $8.0$) and
independent Gaussian noise (scale $0.1$) added to each observation. Each class
has a distinct mean function parameterized using a 5-dimensional, order-3
B-spline. The first class has a declining mean trajectory, the second has a
trajectory that declines then stabilizes, and the third has a stable
trajectory.\footnote{The exact B-spline coefficients can be found in the
  simulation code included in the supplement.} All classes are equally likely
\emph{a priori}. At each measurement time, the treatment policy $\pi$ determines
a probability $p$ of treatment administration (we use only a single treatment
type). The treatments increase the severity marker by a constant amount for 2
hours. If two or more actions occur within 2 hours of one another, the effects
do not add up (i.e. it is as though only one treatment is active). Additional
details about the simulator and policies can be found in the supplement.

Policies $\pi_A$ and $\pi_B$ determine a probability of treatment at each
outcome measurement time. They each use the average of the observed outcomes
over the previous two hours, which we denote using $\hat{y}_{(t-2):t}$, as a
feature, which is then multiplied by a weight $w_A = -0.5$ ($w_B = 0.5$ for
regime $B$) and passed through the inverse logit to determine a probabilty. The
policy $\pi_C$ for regime $C$ depends on the patient's latent class. The
probability of treatment at any time $t$ is $p = \alpha_z \sigma(w_A \cdot
\hat{y}_{(t-2):t})$, where $\alpha_z \in (0, 1)$ is a weight that depends on the
latent class $z$. We set $\alpha_1 = 0.2$, $\alpha_2 = 0.9$, and $\alpha_3 =
0.5$.

\section{Mixture Estimation Details}

For both the simulated and real data experiments, we analytically sum over the
component-specific densities to obtain an explicit mixture density involving no
latent variables. We then estimate the parameters using maximum likelihood. The
likelihood surface is highly non-convex. To account for this, we used different
parameter initialization strategies for the simulated and real data.

On the simulated data experiments, the mixture components for both the CGP and
baseline GP are primarily distinguished by the mean functions. We initialize the
mean parameters for both the baseline GP and CGP by first fitting a linear mixed
model with B-spline bases using the EM algorithm, computing MAP estimates of
trace-specific coefficients, clustering the coefficients, and initializing with
the cluster centers.

On the real data, traces have similar mean behavior (trajectories drift around
the initial creatinine value), but differed by length and amplitude of
variations from the mean. We therefore centered each trace around its initial
creatinine measurement (which we condition on), and use a mean function that
includes only the short-term and long-term response functions. For each mixture,
the response function parameters are initialized randomly: parameters $a$, $b$,
and $r$ are initialized using a $\text{LogNormal}(\text{mean}=0.0,
\text{std}=0.1)$; heights $h_1$ and $h_2$ are initialized using a
$\text{Normal}(\text{mean}=0.0, \text{std}=0.1)$. For each mixture, $\Sigma$
(L300) is initialized to the identity matrix; $\alpha$ and $\nu$ are drawn from
a $\text{LogNormal}(\text{mean}=0.0, \text{std}=0.1)$.

\bibliographystyle{plainnat}
\bibliography{main}

\begin{thebibliography}{46}
\providecommand{\natexlab}[1]{#1}
\providecommand{\url}[1]{\texttt{#1}}
\expandafter\ifx\csname urlstyle\endcsname\relax
  \providecommand{\doi}[1]{doi: #1}\else
  \providecommand{\doi}{doi: \begingroup \urlstyle{rm}\Url}\fi

\bibitem[Alaa et~al.(2016)Alaa, Yoon, Hu, and van~der
  Schaar]{alaa2016personalized}
A.M. Alaa, J.~Yoon, S.~Hu, and M.~van~der Schaar.
\newblock {Personalized Risk Scoring for Critical Care Patients using Mixtures
  of Gaussian Process Experts}.
\newblock In \emph{ICML Workshop on Computational Frameworks for
  Personalization}, 2016.

\bibitem[Arjas and Parner(2004)]{arjas2004causal}
E.~Arjas and J.~Parner.
\newblock Causal reasoning from longitudinal data.
\newblock \emph{Scandinavian Journal of Statistics}, 31\penalty0 (2):\penalty0
  171--187, 2004.

\bibitem[Bottou et~al.(2013)Bottou, Peters, Candela, Charles, Chickering,
  Portugaly, Ray, Simard, and Snelson]{bottou2013counterfactual}
L.~Bottou, J.~Peters, J.Q. Candela, D.X. Charles, M.~Chickering, E.~Portugaly,
  D.~Ray, P.Y. Simard, and E.~Snelson.
\newblock Counterfactual reasoning and learning systems: the example of
  computational advertising.
\newblock \emph{Journal of Machine Learning Research (JMLR)}, 14\penalty0
  (1):\penalty0 3207--3260, 2013.

\bibitem[Brodersen et~al.(2015)Brodersen, Gallusser, Koehler, Remy, and
  Scott]{brodersen2015inferring}
K.H. Brodersen, F.~Gallusser, J.~Koehler, N.~Remy, and S.L. Scott.
\newblock Inferring causal impact using bayesian structural time-series models.
\newblock \emph{The Annals of Applied Statistics}, 9\penalty0 (1):\penalty0
  247--274, 2015.

\bibitem[Caruana et~al.(2015)Caruana, Lou, Gehrke, Koch, Sturm, and
  Elhadad]{caruana2015intelligible}
R.~Caruana, Y.~Lou, J.~Gehrke, P.~Koch, M.~Sturm, and N.~Elhadad.
\newblock Intelligible models for healthcare: Predicting pneumonia risk and
  hospital 30-day readmission.
\newblock In \emph{International Conference on Knowledge Discovery and Data
  Mining (KDD)}, pages 1721--1730. ACM, 2015.

\bibitem[Cheng et~al.(2017)Cheng, Darnell, Chivers, Draugelis, Li, and
  Engelhardt]{cheng2017sparse}
L.F. Cheng, G.~Darnell, C.~Chivers, M.E. Draugelis, K.~Li, and B.E. Engelhardt.
\newblock Sparse multi-output {Gaussian} processes for medical time series
  prediction.
\newblock \emph{arXiv preprint arXiv:1703.09112}, 2017.

\bibitem[Cunningham et~al.(2012)Cunningham, Ghahramani, and
  Rasmussen]{cunningham2012gaussian}
J.~Cunningham, Z.~Ghahramani, and C.E. Rasmussen.
\newblock {Gaussian} processes for time-marked time-series data.
\newblock In \emph{International Conference on Artificial Intelligence and
  Statistics (AISTATS)}, pages 255--263, 2012.

\bibitem[Daley and Vere-Jones(2007)]{daley2007introduction}
D.J. Daley and D.~Vere-Jones.
\newblock \emph{An Introduction to the Theory of Point Processes}.
\newblock Springer Science \& Business Media, 2007.

\bibitem[Doroudi et~al.(2017)Doroudi, Thomas, and
  Brunskill]{doroudi2017importance}
S.~Doroudi, P.S. Thomas, and E.~Brunskill.
\newblock Importance sampling for fair policy selection.
\newblock In \emph{Uncertainty in Artificial Intelligence (UAI)}, 2017.

\bibitem[Dud{\'\i}k et~al.(2011)Dud{\'\i}k, Langford, and Li]{dudik2011doubly}
M.~Dud{\'\i}k, J.~Langford, and L.~Li.
\newblock Doubly robust policy evaluation and learning.
\newblock In \emph{International Conference on Machine Learning (ICML)}, 2011.

\bibitem[Dyagilev and Saria(2016)]{dyagilev2016learning}
K.~Dyagilev and S.~Saria.
\newblock Learning (predictive) risk scores in the presence of censoring due to
  interventions.
\newblock \emph{Machine Learning}, 102\penalty0 (3):\penalty0 323--348, 2016.

\bibitem[Gong et~al.(2016)Gong, Zhang, Liu, Tao, Glymour, and
  Sch{\"o}lkopf]{gong2016domain}
M.~Gong, K.~Zhang, T.~Liu, D.~Tao, C.~Glymour, and B.~Sch{\"o}lkopf.
\newblock Domain adaptation with conditional transferable components.
\newblock In \emph{International Conference on Machine Learning (ICML)}, 2016.

\bibitem[Hawkes(1971)]{hawkes1971spectra}
A.G. Hawkes.
\newblock Spectra of some self-exciting and mutually exciting point processes.
\newblock \emph{Biometrika}, pages 83--90, 1971.

\bibitem[Jiang and Li(2016)]{jiang2016doubly}
N.~Jiang and L.~Li.
\newblock Doubly robust off-policy value evaluation for reinforcement learning.
\newblock In \emph{International Conference on Machine Learning (ICML)}, pages
  652--661, 2016.

\bibitem[Johansson et~al.(2016)Johansson, Shalit, and
  Sontag]{johansson2016learning}
F.D. Johansson, U.~Shalit, and D.~Sontag.
\newblock Learning representations for counterfactual inference.
\newblock In \emph{International Conference on Machine Learning (ICML)}, 2016.

\bibitem[Li-wei et~al.(2015)Li-wei, Adams, Mayaud, Moody, Malhotra, Mark, and
  Nemati]{li2015physiological}
H.L Li-wei, R.P. Adams, L.~Mayaud, G.B. Moody, A.~Malhotra, R.G. Mark, and
  S.~Nemati.
\newblock A physiological time series dynamics-based approach to patient
  monitoring and outcome prediction.
\newblock \emph{IEEE Journal of Biomedical and Health Informatics}, 19\penalty0
  (3):\penalty0 1068--1076, 2015.

\bibitem[Lok(2008)]{lok2008statistical}
J.J. Lok.
\newblock Statistical modeling of causal effects in continuous time.
\newblock \emph{The Annals of Statistics}, pages 1464--1507, 2008.

\bibitem[Mooij et~al.(2013)Mooij, Janzing, and Sch{\"o}lkopf]{mooij2013from}
J.M. Mooij, D.~Janzing, and B.~Sch{\"o}lkopf.
\newblock From ordinary differential equations to structural causal models: the
  deterministic case.
\newblock 2013.

\bibitem[Morgan and Winship(2014)]{morgan2014counterfactuals}
S.L. Morgan and C.~Winship.
\newblock \emph{Counterfactuals and causal inference}.
\newblock Cambridge University Press, 2014.

\bibitem[Murphy(2003)]{murphy2003optimal}
S.A. Murphy.
\newblock Optimal dynamic treatment regimes.
\newblock \emph{Journal of the Royal Statistical Society: Series B (Statistical
  Methodology)}, 65\penalty0 (2):\penalty0 331--355, 2003.

\bibitem[Nahum-Shani et~al.(2012)Nahum-Shani, Qian, Almirall, Pelham, Gnagy,
  Fabiano, Waxmonsky, Yu, and Murphy]{nahum2012q}
I.~Nahum-Shani, M.~Qian, D.~Almirall, W.E. Pelham, B.~Gnagy, G.A. Fabiano, J.G.
  Waxmonsky, J.~Yu, and S.A. Murphy.
\newblock Q-learning: A data analysis method for constructing adaptive
  interventions.
\newblock \emph{Psychological Methods}, 17\penalty0 (4):\penalty0 478, 2012.

\bibitem[Neyman(1923)]{neyman1923applications}
J.~Neyman.
\newblock Sur les applications de la th{\'e}orie des probabilit{\'e}s aux
  experiences agricoles: Essai des principes.
\newblock \emph{Roczniki Nauk Rolniczych}, 10:\penalty0 1--51, 1923.

\bibitem[Neyman(1990)]{neyman1990application}
J.~Neyman.
\newblock On the application of probability theory to agricultural experiments.
\newblock \emph{Statistical Science}, 5\penalty0 (4):\penalty0 465--472, 1990.

\bibitem[Ng et~al.(2006)Ng, Coates, Diel, Ganapathi, Schulte, Tse, Berger, and
  Liang]{ng2006autonomous}
A.Y. Ng, A.~Coates, M.~Diel, V.~Ganapathi, J.~Schulte, B.~Tse, E.~Berger, and
  E.~Liang.
\newblock Autonomous inverted helicopter flight via reinforcement learning.
\newblock In \emph{Experimental Robotics IX}, pages 363--372. Springer, 2006.

\bibitem[Nocedal and Wright(2006)]{nocedal2006numerical}
J.~Nocedal and S.J. Wright.
\newblock Numerical optimization 2nd, 2006.

\bibitem[P{\u{a}}duraru et~al.(2012)P{\u{a}}duraru, Precup, Pineau, and
  Com{\u{a}}nici]{puaduraru2012empirical}
C.~P{\u{a}}duraru, D.~Precup, J.~Pineau, and G.~Com{\u{a}}nici.
\newblock An empirical analysis of off-policy learning in discrete mdps.
\newblock In \emph{Workshop on Reinforcement Learning}, page~89, 2012.

\bibitem[Pearl(2009)]{pearl2009causality}
J.~Pearl.
\newblock \emph{Causality: models, reasoning and inference}.
\newblock Cambridge University Press, 2009.

\bibitem[Rasmussen and Williams(2006)]{rasmussen2006gaussian}
C.E. Rasmussen and C.K.I. Williams.
\newblock \emph{Gaussian processes for machine learning}.
\newblock the MIT Press, 2006.

\bibitem[Ribeiro et~al.(2016)Ribeiro, Singh, and Guestrin]{ribeiro2016should}
M.T. Ribeiro, S.~Singh, and C.~Guestrin.
\newblock Why should i trust you?: Explaining the predictions of any
  classifier.
\newblock In \emph{International Conference on Knowledge Discovery and Data
  Mining (KDD)}, pages 1135--1144. ACM, 2016.

\bibitem[Robins(1986)]{robins1986new}
J.M. Robins.
\newblock A new approach to causal inference in mortality studies with a
  sustained exposure period—application to control of the healthy worker
  survivor effect.
\newblock \emph{Mathematical Modelling}, 7\penalty0 (9-12):\penalty0
  1393--1512, 1986.

\bibitem[Robins(1992)]{robins1992estimation}
J.M. Robins.
\newblock Estimation of the time-dependent accelerated failure time model in
  the presence of confounding factors.
\newblock \emph{Biometrika}, 79\penalty0 (2):\penalty0 321--334, 1992.

\bibitem[Robins(1997)]{robins1997causal}
J.M. Robins.
\newblock Causal inference from complex longitudinal data.
\newblock In \emph{Latent variable modeling and applications to causality},
  pages 69--117. Springer, 1997.

\bibitem[Robins and Hern{\'a}n(2009)]{robins2009estimation}
J.M. Robins and M.A. Hern{\'a}n.
\newblock Estimation of the causal effects of time-varying exposures.
\newblock \emph{Longitudinal data analysis}, pages 553--599, 2009.

\bibitem[Robins et~al.(2000)Robins, Rotnitzky, and
  Scharfstein]{robins2000sensitivity}
J.M. Robins, A.~Rotnitzky, and D.O. Scharfstein.
\newblock Sensitivity analysis for selection bias and unmeasured confounding in
  missing data and causal inference models.
\newblock In \emph{Statistical models in epidemiology, the environment, and
  clinical trials}, pages 1--94. Springer, 2000.

\bibitem[Rubin(1978)]{rubin1978bayesian}
D.B. Rubin.
\newblock Bayesian inference for causal effects: The role of randomization.
\newblock \emph{The Annals of statistics}, pages 34--58, 1978.

\bibitem[Saeed et~al.(2011)Saeed, Villarroel, Reisner, Clifford, Lehman, Moody,
  Heldt, Kyaw, Moody, and Mark]{saeed2011multiparameter}
M.~Saeed, M.~Villarroel, A.T. Reisner, G.~Clifford, L.W. Lehman, G.~Moody,
  T.~Heldt, T.H. Kyaw, B.~Moody, and R.G. Mark.
\newblock Multiparameter intelligent monitoring in intensive care {II}
  ({MIMIC-II}): a public-access intensive care unit database.
\newblock \emph{Critical Care Medicine}, 39\penalty0 (5):\penalty0 952, 2011.

\bibitem[Scharfstein et~al.(2014)Scharfstein, McDermott, Olson, and
  Wiegand]{scharfstein2014global}
D.~Scharfstein, A.~McDermott, W.~Olson, and F.~Wiegand.
\newblock Global sensitivity analysis for repeated measures studies with
  informative dropout: A fully parametric approach.
\newblock \emph{Statistics in Biopharmaceutical Research}, 6\penalty0
  (4):\penalty0 338--348, 2014.

\bibitem[Schulam and Saria(2015)]{schulam2015framework}
P.~Schulam and S.~Saria.
\newblock A framework for individualizing predictions of disease trajectories
  by exploiting multi-resolution structure.
\newblock In \emph{Advances in Neural Information Processing Systems (NIPS)},
  pages 748--756, 2015.

\bibitem[Sokol and Hansen(2014)]{sokol2014causal}
A.~Sokol and N.R. Hansen.
\newblock Causal interpretation of stochastic differential equations.
\newblock \emph{Electronic Journal of Probability}, 19\penalty0 (100):\penalty0
  1--24, 2014.

\bibitem[Soleimani et~al.(2017)Soleimani, Subbaswamy, and
  Saria]{soleimani2017treatment}
H.~Soleimani, A.~Subbaswamy, and S.~Saria.
\newblock Treatment-response models for counterfactual reasoning with
  continuous-time, continuous-valued interventions.
\newblock In \emph{Uncertainty in Artificial Intelligence (UAI)}, 2017.

\bibitem[Sutton and Barto(1998)]{sutton1998reinforcement}
R.S. Sutton and A.G. Barto.
\newblock \emph{Reinforcement learning: An introduction}, volume~1.
\newblock MIT press Cambridge, 1998.

\bibitem[Swaminathan and Joachims(2015)]{swaminathan2015counterfactual}
A.~Swaminathan and T.~Joachims.
\newblock Counterfactual risk minimization.
\newblock In \emph{International Conference on Machine Learning (ICML)}, 2015.

\bibitem[Taubman et~al.(2009)Taubman, Robins, Mittleman, and
  Hern{\'a}n]{taubman2009intervening}
S.L. Taubman, J.M. Robins, M.A. Mittleman, and M.A. Hern{\'a}n.
\newblock Intervening on risk factors for coronary heart disease: an
  application of the parametric g-formula.
\newblock \emph{International Journal of Epidemiology}, 38\penalty0
  (6):\penalty0 1599--1611, 2009.

\bibitem[Taylor et~al.(1994)Taylor, Cumberland, and Sy]{taylor1994stochastic}
J.~Taylor, W.~Cumberland, and J.~Sy.
\newblock A stochastic model for analysis of longitudinal {AIDS} data.
\newblock \emph{Journal of the American Statistical Association}, 89\penalty0
  (427):\penalty0 727--736, 1994.

\bibitem[Wiens et~al.(2016)Wiens, Guttag, and Horvitz]{wiens2016patient}
J.~Wiens, J.~Guttag, and E.~Horvitz.
\newblock Patient risk stratification with time-varying parameters: a multitask
  learning approach.
\newblock \emph{Journal of Machine Learning Research (JMLR)}, 17\penalty0
  (209):\penalty0 1--23, 2016.

\bibitem[Xu et~al.(2016)Xu, Xu, and Saria]{xu2016mlhc}
Y.~Xu, Y.~Xu, and S.~Saria.
\newblock A {Bayesian} nonparametric approach for estimating individualized
  treatment-response curves.
\newblock In \emph{Machine Learning for Healthcare Conference (MLHC)}, pages
  282--300, 2016.

\end{thebibliography}

\end{document}